# TJ4DRadSet: A 4D Radar Dataset for Autonomous Driving

Lianqing Zheng[1], Zhixiong Ma[1,*], Xichan Zhu[1], Bin Tan[1], Sen Li[1], Kai Long[1], Weiqi Sun[1], Sihan Chen[1], Lu Zhang[1], Mengyue Wan[1], Libo Huang[2], Jie Bai[2,*]

*Abstract*— The next-generation high-resolution automotive radar (4D radar) can provide additional elevation measurement and denser point clouds, which has great potential for 3D sensing in autonomous driving. In this paper, we introduce a dataset named TJ4DRadSet with 4D radar points for autonomous driving research. The dataset was collected in various driving scenarios, with a total of 7757 synchronized frames in 44 consecutive sequences, which are well annotated with 3D bounding boxes and track ids. We provide a 4D radar-based 3D object detection baseline for our dataset to demonstrate the effectiveness of deep learning methods for 4D radar point clouds. The dataset can be accessed via the following link: https://github.com/TJRadarLab/TJ4DRadSet.

## I. INTRODUCTION

Autonomous driving technology [1] has recently received much attention. The high-level autonomous driving system mainly consists of modules such as environment perception, road planning, and decision execution [2]. A highly reliable, low-cost, high-resolution perception module is necessary for self-driving vehicles. At the current stage, the perception module mainly uses sensors such as cameras, lidars, and automotive radars to obtain environmental information of different modes [3]. Undeniably, the camera and lidar are vulnerable to harsh conditions such as rain, fog, and intense light, whose performance will decline significantly with the increase of adversity. In contrast, the automotive radar is essential because of its strong robustness [4] and cost-effectiveness. Due to low azimuthal resolution, conventional automotive radar is only used for blind-spot detection, collision warning, and other driving assistance applications. The emergence of new-generation 4D radar [5] makes up for the low definition of conventional automotive radar and provides elevation measurement, which is well suited for applications in high-level autonomous driving. The four dimensions of 4D radar are range, azimuth, elevation, and Doppler velocity. It also provides some other low-level features such as radar-cross-section (RCS) or signal-to-noise ratio (SNR).

3D object detection and tracking are essential to environment perception. With the development of deep learning and artificial intelligence, an enormous amount of neural networks have been applied to 3D perception [6]. Training a 3D object detection network requires large-scale data, which should cover many diverse and complex conditions. What's more, the ground truth of data needs to be accurate for supervised learning to ensure the trained network is valid. Compared with camera and lidar, few autonomous driving datasets contain 4D radar, which limits the research and application of deep learning in the 4D radar point cloud. To fill this gap, we proposed a 4D radar dataset for autonomous driving called TJ4DRadSet. The data collection platform contains multi-sensors, including 4D radar, camera, lidar, and Global Navigation Satellite System (GNSS), as shown in Figure 1. We hope the dataset will facilitate the research of 4D radar-based perception algorithms. Our contributions are listed as follows:

- We present a dataset named TJ4DRadSet, an autonomous driving dataset containing 4D radar point clouds in continuous sequences with 3D annotations, which also provides multi-modal complete information on lidar, camera and GNSS.

- TJ4DRadSet contains 40K frames of synchronized data, where 7757 frames, 44 sequences with high-quality annotated 3D bounding boxes and track ids. The 3D annotation system uses joint multi-sensor annotation and multi-round manual checks.

- TJ4DRadSet covers various road conditions, such as elevated roads, complex intersections, one-way roads, and urban roads. It also includes bad lighting conditions such as intense light and darkness. The dataset is suitable for developing 3D perception algorithms based on the 4D radar to facilitate its application in high-level autonomous driving.

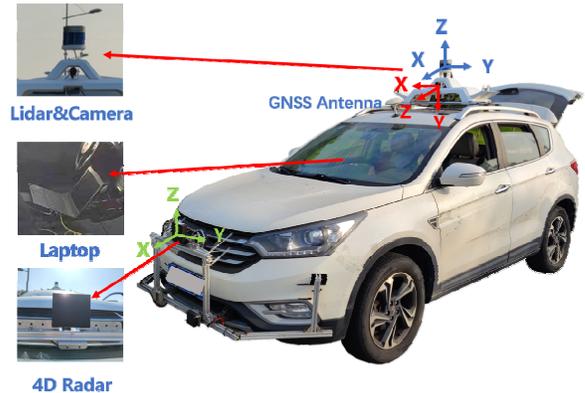

Figure 1.  Data acquisition platform and coordinate system.

[1]Lianqing Zheng, Bin Tan, Sen Li, {zhenglianqing, tanbin, lisen}@tongji.edu.cn

TABLE I. CURRENT DRIVING RADAR DATASETS

| Dataset | Size | Radar Type | Other Modalities | 4D Radar Point Cloud | Object Detection | Object Tracking | 3D Annotations |
|---|---|---|---|---|---|---|---|
| nuScenes[9] | Large | Low Resolution | Lidar&Camera | ✗ | ✓ | ✓ | ✓ |
| RADIATE[12] | Middle | Scanning | Lidar&Camera | ✗ | ✓ | ✓ | ✗ |
| MulRan[13] | Middle | Scanning | Lidar | ✗ | ✗ | ✗ | ✗ |
| RadarScenes[10] | Large | Low Resolution | Camera | ✗ | ✓ | ✓ | ✗ |
| RadarRobotCar[11] | Large | Scanning | Lidar&Camera | ✗ | ✗ | ✗ | ✗ |
| Astyx[14] | Small | High Resolution | Lidar&Camera | ✓ | ✓ | ✗ | ✓ |
| RADIal[15] | Middle | High Resolution | Lidar&Camera | ✓ | ✓ | ✗ | ✗ |
| VoD[16] | Middle | High Resolution | Lidar&Camera | ✓ | ✓ | ✓ | ✓ |
| **TJ4DRadSet(ours)** | Middle | High Resolution | Lidar&Camera | ✓ | ✓ | ✓ | ✓ |

- Based on TJ4DRadSet, we provide a baseline for 4D radar-based 3D object detection. The results show that 4D radar has a promising potential for high-level autonomous driving.

The paper is organized as follows: Section II introduces related work on other datasets. Section III describes our dataset in detail. In Section IV, we perform the baseline result of 3D object detection based on 4D radar. A brief conclusion and future work are presented in Section V.

## II. RELATED WORK

Deep learning technique is playing an increasing role in autonomous driving. It relies on a large amount of high-quality data. Therefore, a growing number of open dataset benchmarks have appeared in recent years, such as KITTI [7] and Waymo Open [8], which have contributed to the advancement of autonomous driving technology. With these benchmarks, we can evaluate the performance of different algorithms for various tasks.

Automotive radar has proven to be an effective sensor due to its robustness in all weather and low price. However, many datasets do not contain radar sensors, which limits the application of data-driven algorithms based on radar data. Since the nuScenes [9] dataset was released, some datasets with radar data started to appear, which has aroused people's interest in radar. The comparison of each dataset containing radar data is shown in TABLE I. Some datasets contain low-resolution FMCW radars, such as nuScenes, and RadarScenes [10], whose radar point clouds lack elevation information for accurate 3D perception. Some datasets use scanning radar to collect data, such as RadarRobotCar [11], RADIATE [12], and MulRan [13], whose radar data are mainly interpreted as image data and lack Doppler velocity. For the new-generation 4D imaging radar, the 4D point cloud will be the primary output format, containing spatial and velocity information. Currently, Astyx [14], RADIal [15] and VoD[16] dataset have high-resolution 4D radar sensor. Astyx has only 545 frames of point cloud data, which is small and lacks tracking information. RADIal contains complete radar formats, such as range-Doppler maps and point clouds, which only has 2D labeled boxes and a "Car" label. VoD dataset is a novel automotive dataset, which is the work of the same period as ours. VoD contains 8600 frames of synchronized and calibrated lidar, camera, and 4D radar data acquired in urban traffic, which also provides 3D annotations and track ids. Compared to the VoD dataset, our dataset contains much richer and more challenging driving scenario clips.

## III. THE TJ4DRADSET DATASET

In this section, we introduce sensor parameters, calibration, data collection and annotation, then provide statistical analysis and visualization.

### A. Sensors

The TJ4DRadSet mainly contains 4D radar, lidar and camera. As shown in Figure 1, the camera and lidar are mounted on the roof bracket, and the 4D radar is installed in the middle of the front ventilation gride. The lidar can do 360-degree scanning of environmental information, while the camera and 4D radar capture the information in the field of view (FOV) ahead, covering the forward driving view. The main parameters of each sensor are shown in TABLE II. In addition, the GNSS information is included and corrected by real-time kinematic (RTK) to achieve high-precision positioning, which has the speed and location information of the ego vehicle.

### B. Sensor Calibration

Multi-sensor calibration is the basis for perception algorithms. The process mainly consists of intrinsic parameters calibration, extrinsic calibration, and temporal alignment. The intrinsic parameters and distortion coefficients of the camera are calibrated by MATLAB Toolkit [17] and a checkerboard. The distortion coefficients are used for correction to obtain rectified images. The intrinsic parameters of 4D radar and lidar have been calibrated offline at the factory.

It can be divided into two processes for extrinsic parameters: camera and lidar extrinsic calibration; 4D radar and lidar extrinsic calibration. The extrinsic parameters of the camera and 4D radar can be obtained by performing matrix operations on the remaining two extrinsic parameters. The extrinsic parameters between the different sensors are represented as translation and rotation matrix. For camera and lidar extrinsic calibration, we used a checkerboard to perform 2D-3D alignment of the point cloud and image data to complete a rough calibration. Then, we manually fine-tune the extrinsic parameters by static objects such as trees and poles in the environment. For 4D radar and lidar extrinsic calibration, we consider it as 3D-3D point cloud alignment in space. Firstly, the distance between the two sensors is measured as a rough translation parameter. Then extrinsic parameters are fine-tuned by using multiple angular reflectors in space.

TABLE II. SPECIFICATION OF THE TJ4DRADSET'S SENSOR SUITES

| Parameters\Sensors | Resolution | | | FOV | | | FPS |
|---|---|---|---|---|---|---|---|
| | Range | Azimuth | Elevation | Range | Azimuth | Elevation | |
| Camera | | 1280px | 960px | | 66.5° | 94° | 30 |
| Lidar | 0.03m | 0.1°-0.4° | 0.33° | 120m | 360° | 40° | 10 |
| 4D Radar | 0.86m | <1° | <1° | 400m | 113° | 45° | 15 |

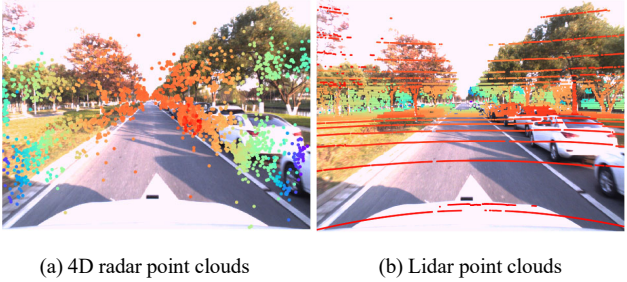

(a) 4D radar point clouds     (b) Lidar point clouds

Figure 2. 4D radar and lidar point clouds projection.

All sensors work under the ROS driver. Since each sensor runs at different frame rates, we align the data by using the arrival time of the data as the timestamp. The final 4D radar and lidar point clouds are projected into the image, as shown in Figure 2.

### C. Data Collection and Annotation

TJ4DRadSet was collected in Suzhou, China, in the fourth quarter of 2021. Figure 3 records the location of the data collection. The dataset covers a wide range of driving conditions, including various lighting conditions, such as normal lighting, bright light and darkness, and different road types, such as urban roads, elevated roads, and industrial zones. There are complex scenarios such as object-dense intersections and simple scenarios such as one-way streets with a few objects. Our acquisition system is based on ROS and all sensor data are recorded in "rosbag" completely.

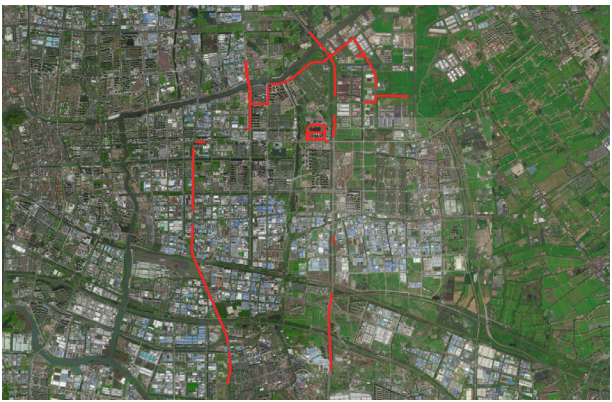

Figure 3. Locations of the data collection.

The ground truth annotation of the dataset mainly includes a 3D bounding box, class and track id for each object. The lidar sensor has a higher point cloud density than 4D radar, which provides a more detailed description of objects' shapes. Therefore, our annotation system mainly relies on lidar point clouds and images for joint annotation. However, some objects that have few lidar points due to occlusion may still appear in 4D radar FOV because of the multipath effect, and we still label them. We finished the annotation manually and reviewed many rounds to ensure the quality of the dataset.

The 3D bounding box of each object includes the center point $(x, y, z)$, length, width, height $(l, w, h)$, and orientation angle (yaw). In addition, we provide occlusion and truncation indicators to distinguish different difficulty levels. The dataset has eight classes (Car, Bus, Truck, Engineering Vehicle, Pedestrian, Motorcyclist, Cyclist, and Tricyclist). In order to have a balanced label distribution and to improve the performance of networks, we map the "Bus" and "Engineering Vehicle" (large) to "Truck", the "Motorcyclist" to "Cyclist". The class of other objects is mapped to "Other Vehicle". The original classes are retained so that the mapping can be customized as needed. We assign a unique id to each object for the tracking task. Finally, 40K frames of synchronized data are extracted, of which 7757 frames in 44 consecutive sequences are labeled.

### D. Dataset Statistics

In this part, we perform some statistical analysis of the dataset. Figure 4(a) shows the number of objects for each class, with "Car" being the most numerous, followed by "Cyclist". The amount of "Truck" and "Pedestrian" is approximately the same. Figure 4(b) shows the speed distribution of the ego vehicle. The distribution of the point cloud density of lidar and 4D radar is shown in Figure 4(c)(d). It can be seen that the 4D radar point cloud is sparser than the laser point cloud, but radar points contain more features, such as Doppler velocity. In addition, we also count the distribution of the distances and orientations of the main classes, shown in Figure 5. Some typical scenarios are visualized in Figure 6.

## IV. BASELINE EXPERIMENTS

This section establishes baselines of 3D object detection based on 4D radar and lidar. We divide the dataset into a training set and test set by sequence and keep the test set with good coverage. In this way, we get 5717 training and 2040 test samples and keep the data split fixed.

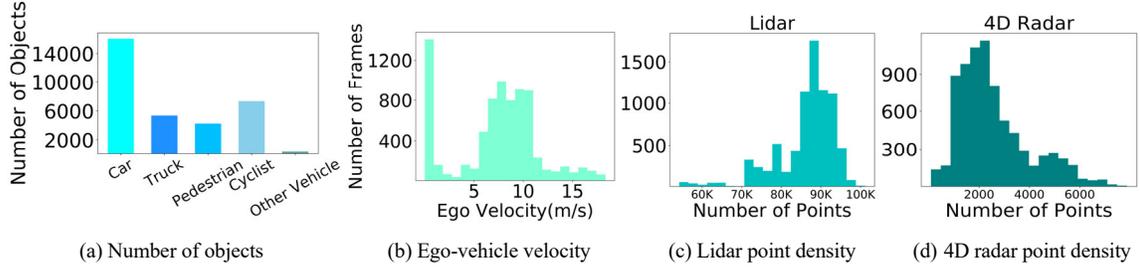

Figure 4. Some basic statistics.

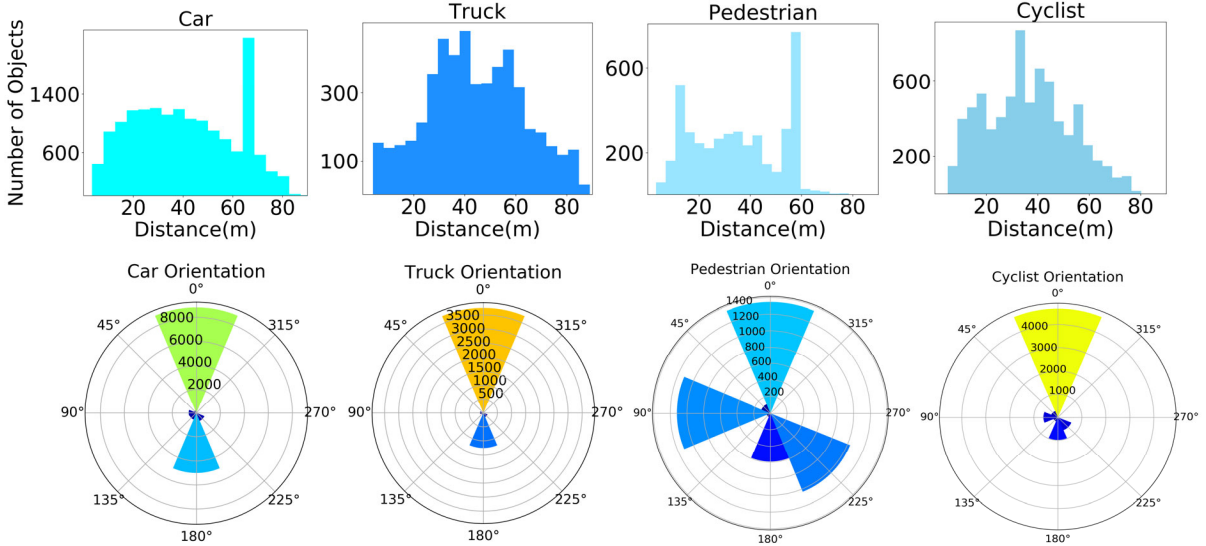

Figure 5. Distribution of the distances and orientations of "Car", "Truck", "Pedestrian" and "Cyclist".

TABLE III. BASELINE RESULTS (4D RADAR)

| Class | BEV-50m | | 3D-50m | | BEV-70m | | 3D-70m | |
|---|---|---|---|---|---|---|---|---|
| | AP@0.5 | AP@0.25 | AP@0.5 | AP@0.25 | AP@0.5 | AP@0.25 | AP@0.5 | AP@0.25 |
| Car | 23.06 | 36.73 | 12.63 | 27.96 | 26.19 | 40.14 | 16.85 | 33.30 |
| Truck | 16.76 | 36.37 | 12.64 | 31.33 | 13.46 | 30.49 | 10.07 | 25.51 |
| Pedestrian | | 35.24 | | 27.64 | | 35.26 | | 27.19 |
| Cyclist | 21.62 | 40.26 | 18.34 | 38.42 | 21.38 | 39.80 | 17.70 | 38.20 |

TABLE IV. BASELINE RESULTS (LIDAR)

| Class | BEV-50m | | 3D-50m | | BEV-70m | | 3D-70m | |
|---|---|---|---|---|---|---|---|---|
| | AP@0.5 | AP@0.25 | AP@0.5 | AP@0.25 | AP@0.5 | AP@0.25 | AP@0.5 | AP@0.25 |
| Car | 69.69 | 69.79 | 69.47 | 69.72 | 52.76 | 52.77 | 52.67 | 52.76 |
| Truck | 37.45 | 49.06 | 30.47 | 43.34 | 24.72 | 33.42 | 22.74 | 31.89 |
| Pedestrian | | 56.28 | | 56.09 | | 49.23 | | 49.07 |
| Cyclist | 53.00 | 54.12 | 49.28 | 54.12 | 44.72 | 48.32 | 43.09 | 48.59 |

The original annotations are under the lidar coordinate system, and we transfer the labels to the 4D radar coordinate system through the lidar-radar extrinsic matrix. Due to the sparsity of radar point clouds, some of the existing networks are difficult to be applied directly to this data format. In this paper, we use PointPillars [18] as the baseline algorithm for 4D radar and lidar because of its good adaptability and the trade-off in both speed and precision. To adapt the 4D radar data, we have partially modified the original configuration and retrained the model using TJ4DRadSet. The detection range along the x-axis is set to 69.12m. We use five-dimensional features of radar point clouds, which include spatial information $(x, y, z)$, Doppler velocity $(v)$ and signal to noise ratio $(s)$. The Doppler velocity $(v)$ is the absolute radial velocity after compensation by ego-motion. In terms of network parameters, we choose the pillar size to be $(0.16m, 0.16m)$. The anchor size format is defined as $(l, w, h)$. For the four classes ("Car", "Truck", "Pedestrian", and "Cyclist"), the anchor sizes are listed as follows: $(4.56m, 1.84m, 1.70m)$, $(10.76m, 2.66m, 3.47m)$, $(0.80m, 0.60m, 1.69m)$, $(1.77m, 0.78m, 1.60m)$. Besides, some data augmentations are used to enhance the robustness of the network, including the world random rotation and random scaling. We use the Adam optimizer [19] to train the model for 80 epochs.

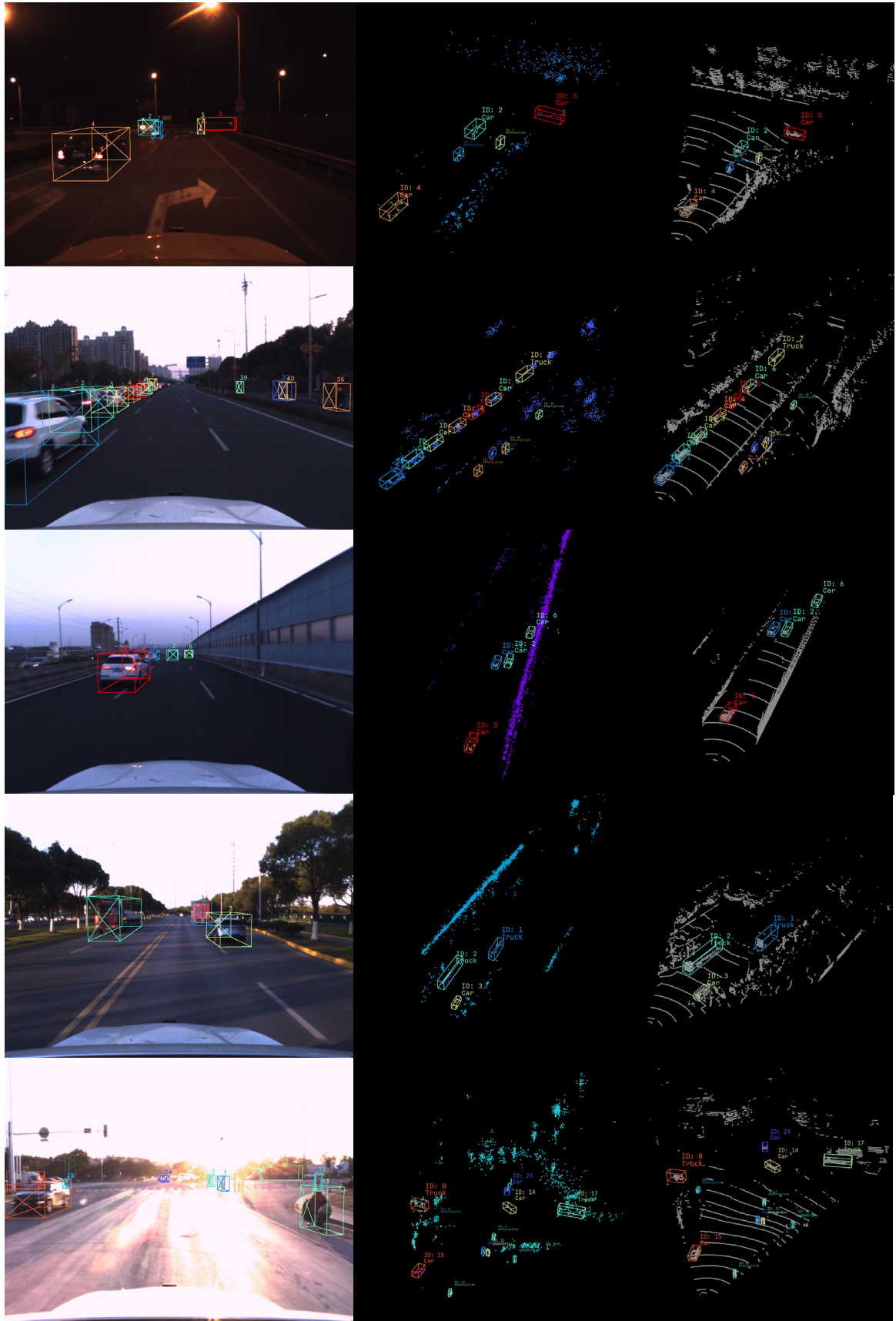

Figure 6. Visualization of typical samples of TJ4DRadSet.

In the evaluation stage, the average precision (AP) is chosen as the metric to evaluate the detection results for each class. Specifically, we use 0.5 and 0.25 IoU thresholds to test "Car", "Truck", and "Cyclist", and only use the 0.25 IoU threshold to evaluate "Pedestrian". We denote the AP under these two thresholds as AP@0.5 and AP@0.25. TABLE III and TABLE IV show the baseline performance at different distances (50m and 70m) and views(BEV, 3D) using 4D radar and lidar, respectively. The results clearly illustrate that 4D radar has potential for 3D perception. In the BEV view, the average accuracy for all classes is over 30% at the 0.25 IoU threshold. Although the baseline algorithm can achieve some results, there is still a big gap between 4D radar and lidar. Under the same algorithm (PointPillars), lidar detection results completely outperform 4D radar, which can be due to several reasons. First, 4D radar has a lower point density, which maybe makes it difficult for the baseline network to extract features effectively. In addition, different data augmentations could also have an impact on the results. It is of great concern how to better extract 4D radar point cloud features and fuse information from other modalities.

## V. Conclusion and Future Work

In this paper, we introduce TJ4DRadSet, a multi-modal autonomous driving dataset containing 4D radar point cloud. The dataset is used to study 4D radar-based 3D perception algorithms. We provide a detailed description of the dataset and conduct baseline experiments. In the future, we will further expand the dataset and research fusion algorithms, point cloud enhancement and feature representation based on 4D radar.